\documentclass[letterpaper,10pt,conference]{ieeeconf}
\IEEEoverridecommandlockouts                              

\usepackage{algorithm}
\usepackage{algorithmic}
\usepackage{array}
\usepackage{amsmath}
\usepackage{amssymb}
\usepackage{booktabs}
\usepackage{color}
\usepackage{dsfont}
\usepackage{epsfig}
\usepackage{graphics}
\usepackage[utf8]{inputenc}
\usepackage{listings}
\usepackage{makecell}
\usepackage{multirow}
\usepackage{pifont}
\usepackage{times}
\usepackage{xcolor}
\usepackage{hyperref}
\usepackage{cleveref}
\title{\LARGE \textbf{Personalized Autonomous Driving with Large Language Models: \\ Field Experiments}}

\author {
    Can~Cui, 
    Zichong~Yang,
    Yupeng~Zhou,
    Yunsheng~Ma,
    Juanwu~Lu,
    Lingxi~Li,
    Yaobin~Chen,
    Jitesh~Panchal,\\
    and~Ziran~Wang
\thanks{C. Cui, Z. Yang, and Y. Zhou contributed equally to this work.
C. Cui, Z. Yang, Y. Zhou, Y. Ma, J. Lu, L. Li, Y. Chen, J. Panchal, and Z. Wang are with the College of Engineering, Purdue University, West Lafayette, IN 47907, USA. Corresponding author: C. Cui, email: {\tt\small cancui@purdue.edu.}}%
}
\begin{document}






\markboth{Journal of \LaTeX\ Class Files,~Vol.~14, No.~8, August~2015}%
{Shell \MakeLowercase{\textit{et al.}}: Bare Demo of IEEEtran.cls for IEEE Journals}

\maketitle%

\begin{abstract}

Integrating large language models (LLMs) in autonomous vehicles enables conversation with AI systems to drive the vehicle. However, it also emphasizes the requirement for such systems to comprehend commands accurately and achieve higher-level personalization to adapt to the preferences of drivers or passengers over a more extended period. In this paper, we introduce an LLM-based framework, \textit{Talk2Drive}, capable of translating natural verbal commands into executable controls and learning to satisfy personal preferences for safety, efficiency, and comfort with a proposed memory module. This is the \textbf{first-of-its-kind} multi-scenario field experiment that deploys LLMs on a real-world autonomous vehicle. Experiments showcase that the proposed system can comprehend human intentions at different intuition levels, ranging from direct commands like ``can you drive faster'' to indirect commands like ``I am really in a hurry now''. Additionally, we use the takeover rate to quantify the trust of human drivers in the LLM-based autonomous driving system, where \textit{Talk2Drive} significantly reduces the takeover rate in highway, intersection, and parking scenarios. We also validate that the proposed memory module considers personalized preferences and further reduces the takeover rate by up to 65.2\% compared with those without a memory module. The experiment video can be watched at \url{https://www.youtube.com/watch?v=4BWsfPaq1Ro}.



\end{abstract}


\section{INTRODUCTION}
\label{sec:intro}

Verbal command understanding has been getting emerging attention in autonomous driving~\cite {chen_driving_2023,cui2023survey}. This requires not only a technical translation and understanding of commands but also a common-sense grasp and an emotional understanding of the nuances inherent in human speech. Consequently, Large Language Models (LLMs) have emerged in popularity among researchers in this field~\cite{wei_chain--thought_2022,kojima_large_2022}. Their comprehensive knowledge base, sourced from numerous data, and strong reasoning abilities enable them to interpret and act upon a wide range of human inputs with remarkable accuracy and human-like understanding.

Traditional approaches in the field of autonomous driving, while effective in certain aspects, confront several critical limitations when it comes to understanding and adapting to the complex commands from humans:
\begin{itemize}
  \item Traditional autonomous driving systems may overlook the importance of providing personalized driving experiences. Many of them do not create profiles from historical configuration data and adjust accordingly to bring human preferences to various scenarios.
  \item Conventional systems struggle to interpret and adapt to the abstract instructions from humans. Humans are good at expressing their feelings on whether the current driving pattern is comfortable to them but it is hard to give commands in exact numerical expressions, where such indirectness poses challenges to traditional autonomous driving systems as they lack deeper contextual and emotional understanding.
  \item Most current autonomous driving systems are trained on limited datasets, which may not cover a wide range of driving scenarios or lack the depth of real-world knowledge required for robust decision-making. As a result, these systems may struggle to make appropriate decisions when faced with unfamiliar or uncommon situations, potentially leading to safety risks.
\end{itemize}

In recent developments, various studies have explored integrating LLMs into autonomous driving. Cui et al. introduced a framework that uses LLMs to engage in the high-level decision-making process~\cite{cui2024drive, cui2024receive}.  GPT-Driver regarded motion planning as a language modeling problem and used LLMs to generate their trajectories and also involved them in the decision-making stage based on the textual descriptions of coordinates~\cite{mao_gpt-driver_2023}. Fu et al. employed LLMs for reasoning and generated actionable driving behaviors~\cite{fu_drive_2023}. Xu et al. used LLMs to answer driving questions for drivers, showing their ability to solve expansibility challenges~\cite{xu_drivegpt4_2023}. DiLu utilized a prompt generator to provide prompts to LLMs and the decision decoder will take actions based on LLMs' responses~\cite{wen_dilu_2023}. Ma et al. proposed an open benchmark dataset, LaMPilot, to facilitate the research of LLMs in autonomous driving~\cite{ma2023lampilot}. However, the current application of LLM-based systems in autonomous driving predominantly centers on simulation environments. Real-world experiments on real vehicles to verify the effectiveness of these systems have not yet been extensively conducted.

There are also studies on systems offering personalized experiences~\cite{wang2022gaussian,zhao2022personalized,ma2024driver}. Du et al. proposed a personalized federated learning framework for predicting lane-change maneuvers by using monitored driver intentions~\cite{du2023driver}. As a particular application of the proposed mobility digital twin framework~\cite{wang2022mobility}, Liao et al. developed a driver digital twin for online prediction of personalized lane change behavior in mixed traffic~\cite{liao2023driver}. However, such personalization frameworks often encounter limitations such as dynamically adapting to human preferences or unseen traffic scenarios. This is where LLMs could potentially complement these systems by offering more nuanced and context-aware adaptations, leveraging their advanced language understanding and generative capabilities.

To tackle the aforementioned limitations, we integrate LLMs into autonomous driving systems and introduce a novel framework known as \textit{Talk2Drive}. In particular, it transforms verbal commands from humans into textual instructions, which are then processed by LLMs on the cloud. Due to their powerful understanding ability, LLMs can interpret and understand humans' intentions precisely. Additionally, LLMs can generate specific driving-related programs that are executed by the autonomous vehicle, adjusting driving behaviors and control parameters to align with the human's preferences. Our work demonstrates a successful end-to-end implementation of an LLM-based autonomous driving system in a real-world vehicle, which is the \textbf{first of its kind} to the best of our knowledge. The main contributions of this paper are summarized as follows:
\begin{itemize}
    \item We integrate our \textit{Talk2Drive} framework into a real vehicle, where a wide range of intelligible command levels are tested with an advanced LLM GPT4~\cite{openai_gpt-4_2023} employed. Our \textit{Talk2Drive} framework can interpret not only the literal meaning of different levels of verbal commands but also their context and the human's emotional state, which allows for a deeper understanding of human needs.
    \item Field experiments are conducted in various scenarios including highway, intersection, and parking, demonstrating that the \textit{Talk2Drive} framework significantly enhances personalization by reducing the driver takeover rate by 75.9\% while maintaining safety and comfort within acceptable levels.
    \item A novel memory module with past interactions is developed to enhance the personalization aspect and allow the vehicle to adapt to individual preferences over time, further reducing the takeover rate by up to 65.2\% compared with those without the memory module.
\end{itemize}




\section{Talk2Drive Framework}
\label{sec:talk2drive_frame}
\begin{figure}[t!]
    \centering
    \includegraphics[width=\linewidth]{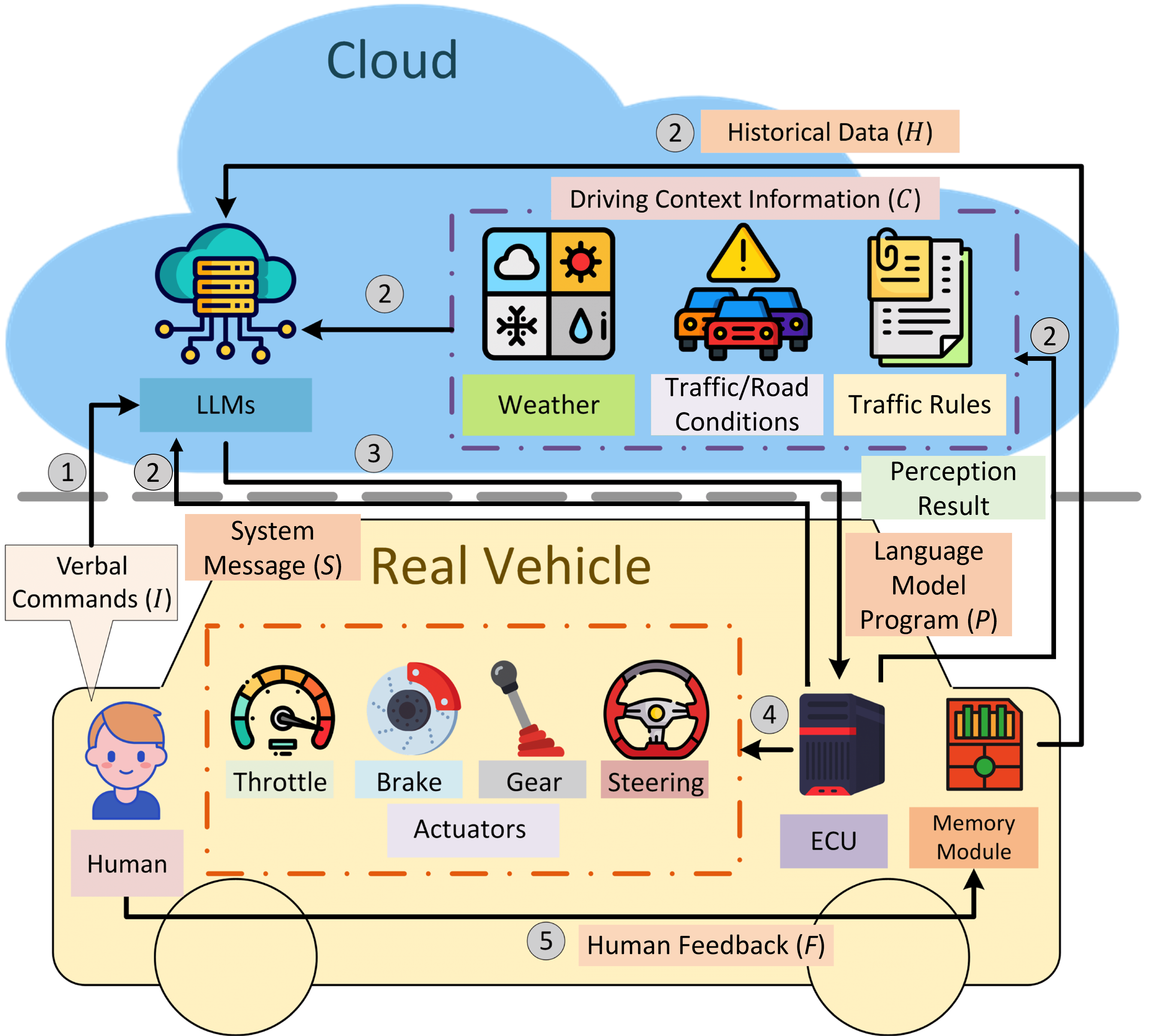}
    \caption{\textit{Talk2Drive} framework architecture. A human's spoken instructions $I$ are processed by cloud-based LLMs, which synthesize contextual data $C$ from weather, traffic conditions, local traffic rules information and the perception results from the local end. Simultaneously, the system message $S$ and the historical data $H$ are sent to LLMs. Then, the LLMs generate executable LMPs $P$ that are communicated to the vehicle's Electronic Control Unit (ECU). These LMPs operate the actuation of vehicle controls, ensuring that the human's intent is translated into safe and personalized driving actions. A memory module archives every command $I$, its resultant LMPs $P$, and subsequent user feedback $F$, ensuring continuous refinement of the personalized driving experience.}
    \vspace{-5mm}
    \label{fig: framework}
\end{figure}

\begin{figure}[t!]
    \centering
    \includegraphics[width=\linewidth]{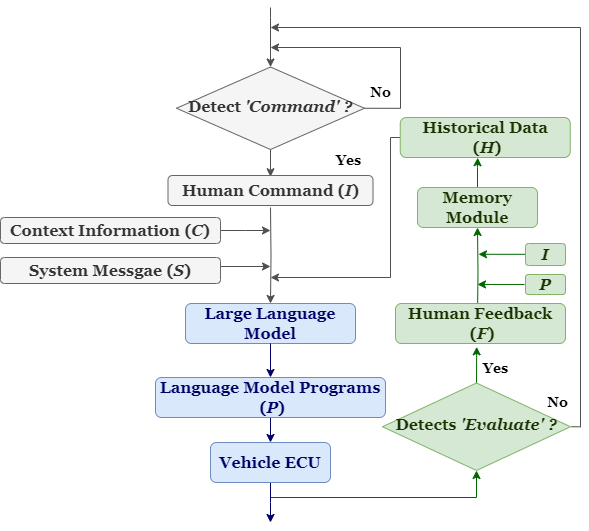}
    \caption{The flowchart of Talk2drive. After the speech recognition module detects the keyword `command', the inputs ($I,C,S,H$) are sent to the LLM. Then, the LLM generates corresponding LMPs to be executed by the ECU. If the speech recognition module detects the keyword 'evaluate', the system receives human feedback ($F$), and both $F$ and its corresponding $I$ and $P$ are updated in the memory module.}
    \vspace{-5mm}
    \label{fig: flowchart}
\end{figure}

This paper proposes \textit{Talk2Drive} (see Fig.~\ref{fig: framework}), an innovative approach to leveraging LLMs to enhance command interpretation and enable personalized decision-making in autonomous vehicles. It integrates cloud-based LLMs to enable personalized comprehension and translation of human commands into executable control sequences with real-time vehicle dynamic inputs. This section begins with a problem statement and then articulates the distinct roles of each cloud-side and vehicle-side operation. The flowchart of the Talk2Drive system is shown in Fig.~\ref{fig: flowchart}.


\subsection{Problem Statement}
In \textit{Talk2Drive}, the model aims to translate verbal commands into executable control sequences for the vehicle. Without losing generality, we denote the verbal commands by $\boldsymbol{I}$ as a string sequence. The cloud-based LLM acts as a translating function $f:\boldsymbol{I}\rightarrow\boldsymbol{P}$ that generates corresponding Language Model Programs (LMPs) as the policy ($\boldsymbol{P}$) for maneuvers.

Besides verbal commands, we consider two additional inputs: contextual data $C$ and system messages $S$. The contextual data helps describe real-time traffic conditions, such as weather, local traffic, and regulations. The system messages help specify tasks and high-level driving logic associated with the current vehicle. In practice, both $C$ and $S$ utilize predefined structured language generators, and the system supplies the appropriate values for $C$ based on the context information.

Nevertheless, for personalization, we need human feedback on execution $F(I,P)$ to evaluate if the generated policy addresses the preferences of the driver or the passengers. We propose an extra memory module (see~\ref{sec:mm}) with evaluation to allow fine-tuning the LLM for learning preferences. Therefore, there are two stages in the \textit{Talk2Drive} workflow:
\begin{equation}
    \begin{aligned}
        \text{Execution}:\quad & P \leftarrow f(I,S,C,H); \\
        \text{Evaluation}:\quad & H \leftarrow \left[I,P,F(I,P)\right].
    \end{aligned}
\end{equation}

\subsection{Command Translation and Contextual Data Integration}  The initial step in the \textit{Talk2Drive} framework involves directly receiving arbitrary verbal commands from humans. Utilizing cutting-edge voice recognition technology, specifically the open-source API Whisper~\cite{radford2023robust}, these verbal commands are accurately captured and then translated into textual instructions ($I$). This translation is crucial for ensuring that the contents and specificities of the human’s spoken words are effectively converted into a textual format that is ready for processing by LLMs. An instance for $I$ is:

\begin{align}
&I \to \label{eq:prompt_command}
\text{Could you drive more conservatively?}
\end{align}

Simultaneously, LLMs access additional cloud-based real-time environment data including weather updates, traffic conditions, and local traffic rules information. For example, LLMs can be empowered by the weather information through Openweather API~\cite{openweather}, the map information (such as road type and speed limits) through OpenStreetMap API~\cite{OpenStreetMap}, and traffic information through TomTom API~\cite{tomtom}. Additionally, the driving context information ($I$) also considers the states of other traffic participants, including their speed and positions. To make all this context data ($C$) accessible to LLMs, we construct an interface that converts the numerical vectors into descriptive text to inform the decision-making process. This approach sets our work apart from other recent advancements in the field~\cite{fu_drive_2023}, where numerical vectors are directly fed into LLMs without any contextual translation. Specifically, we use a predefined structured language generator, and our system will supply the necessary values (the \textcolor{red}{red} contents) to this generator, as shown below:

\begin{align}
&C \to \label{eq:prompt} \\
&\left\{\begin{array}{l}
\text{A vehicle in front of you is running at \textcolor{red}{38.0} km/h.}\\
\text{Your current speed is \textcolor{red}{40.0} km/h.} \\
\text{The speed limit is \textcolor{red}{60.0} km/h.} \\
\text{The weather is \textcolor{red}{sunny}.} \\
\vdots
\end{array}\right. \nonumber
\end{align}

The function translates the state vector's numerical data into a narrative form, allowing the LLM to comprehend the information without requiring additional fine-tuning.

\subsection{Processing and Reasoning with LLMs} 

The next step is to process and reason the textual commands with LLMs. It is important to note that our LLMs are trained using in-context learning, coupled with chain-of-thought prompting. This method has been proven to enhance reasoning ability in language models. The chain-of-thought prompting involves providing a few chain-of-thought demonstrations as exemplars within the prompts to enhance performance on reasoning. Specifically, A chain of thought is a series of textual reasoning steps that lead to the final output and the prompt consists of triples: $\{input,thought, output\}$. An example can be found in \ref{eq:syste_mess}. 

Once a command is translated into texts, it is uploaded to LLMs hosted on the cloud. In our experiments, we utilize GPT-4~\cite{openai_gpt-4_2023} as our LLM and access it through ChatGPT API. This is where the core of \textit{Talk2Drive}'s functionality lies. The LLMs engage in a reasoning process to interpret the command ($I$). Furthermore, the LLMs incorporate contextual data ($C$) provided in the last step. The information will be transferred as input prompts to LLMs. For instance, if the current weather condition indicates snow, the LLMs can intelligently know that initiating movement at a lower speed would be safer. Simultaneously, the system messages ($S$), which define the task and provide high-level driving logic for the entire system using chain-of-thought prompting, are sent to the LLMs.  A simplified example is shown in Eq. \ref{eq:syste_mess}. Additionally, the LLMs also access the memory module, a repository of historical interactions ($H$), to consider the human's past behaviors and preferences. More details about the memory module are in Sec. \ref{sec:mm}.

\begin{align}
&S \to \label{eq:syste_mess} \\
&\left\{\begin{array}{l}
\text{You are an autonomous vehicle with Adaptive cruise}\\
\text{control (ACC) and Lane Keeping Assist (LKA) always}\\
\text{enabled.} \\
\text{You are using Pure Pursuit Controller to do the} \\
\text{waypoint following.} \\
\vdots\\
\text{Here are some examples of how you need to react.}\\
\text{Query: You drive too aggressively.}\\
\text{Thought: The drivers think I drive too fast which looks}\\
\text{aggressive and the drivers do not ask me to change}\\
\text{lanes, so I need to slow down my speed.}\\
\text{Action: ...}\\
\vdots
\end{array}\right. \nonumber
\end{align}

\subsection{Actionable Code Generation} Inspired by the concept of ``Code as Policy~\cite{liang_code_2023}," then we utilize LLMs to generate corresponding LMPs ($P$) based on this interpretation, as these LMP-based policies can efficiently adjust policy code and parameters to accommodate novel tasks and behaviors specified by natural language instructions that have not been seen before. These LMPs are not just simple directives; they include complex driving behaviors and parameters that need to be adjusted in the vehicle’s high-level controllers. Specifically, the LMPs adjust control parameters like the look-ahead distance and look-ahead ratio to optimize pure pursuit~\cite{pure_pursuit} performance. Additionally, LMPs also modify the target velocity of the vehicle to meet drivers' commands. These LMPs take the form of ROS topic commands, directing the autonomous driving system based on Autoware \cite{autoware} to modify its trajectory following configuration.


One simple example of LMP is shown as follows:
\begin{figure}[t!]
    \centering
    \includegraphics[width=\linewidth]{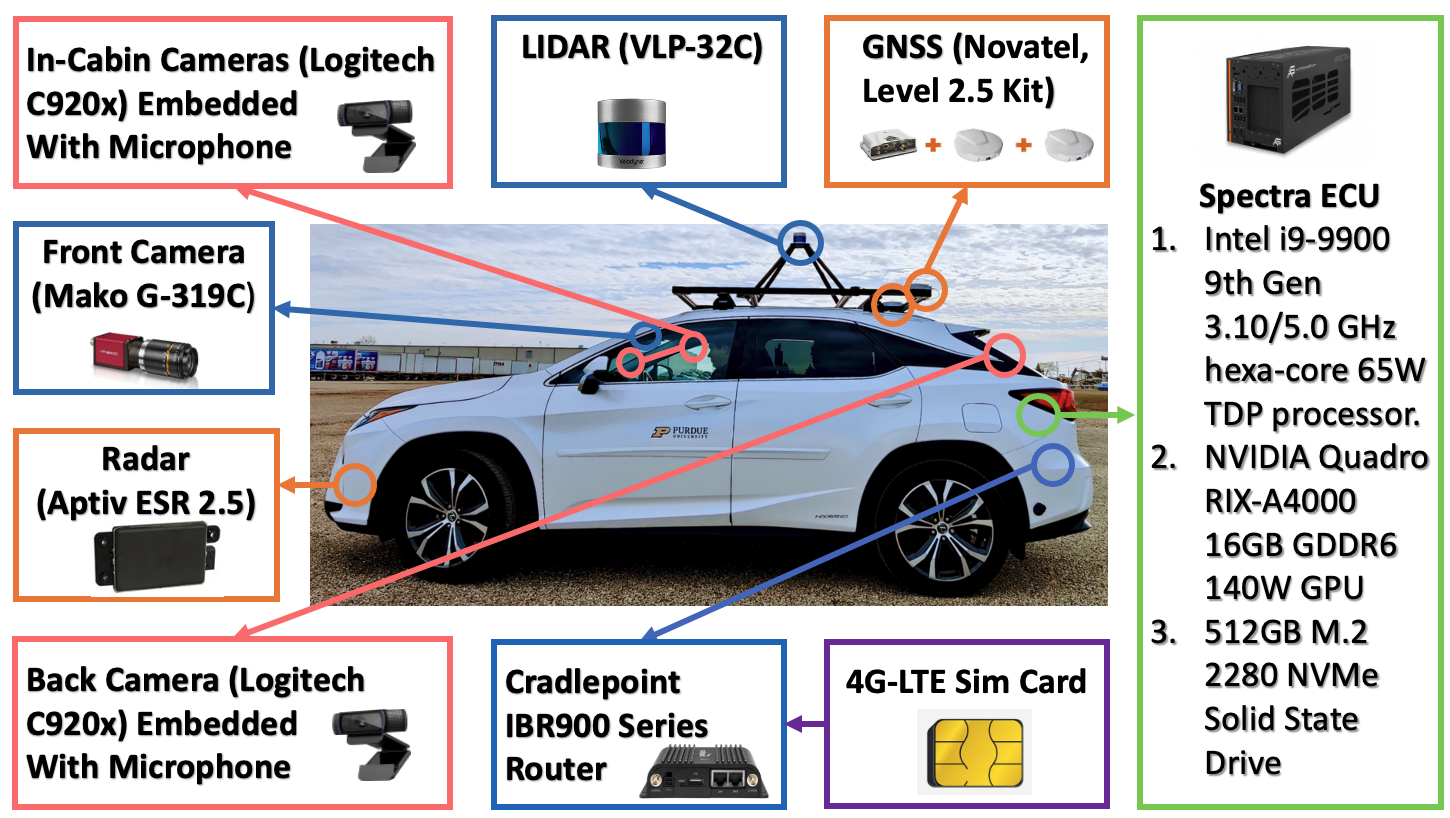}
    \caption{Setup of the autonomous vehicle in the experiment.}
    \vspace{-5mm}
    \label{fig:Vehicle_Setup}
\end{figure}

\lstset{basicstyle=\ttfamily,
  showstringspaces=false,
  commentstyle=\color{red},
  keywordstyle=\color{blue}
}
\begin{flalign}
     P \to && 
  \end{flalign}
\begin{lstlisting}[language=bash]
  $ timeout 1s rostopic pub /vehicle/engage 
  std_msgs/Bool "data: true"
  $ rostopic pub /autoware_config_msgs
  /ConfigWaypointFollower 
  "{\"param_flag\": 1, \"velocity\": 40, 
  \"lookahead_distance\": 12, 
  \"lookahead_ratio\": 2.0}"
  ...
\end{lstlisting}


\begin{figure*}[!t]
    \centering
    \includegraphics[width=\textwidth]{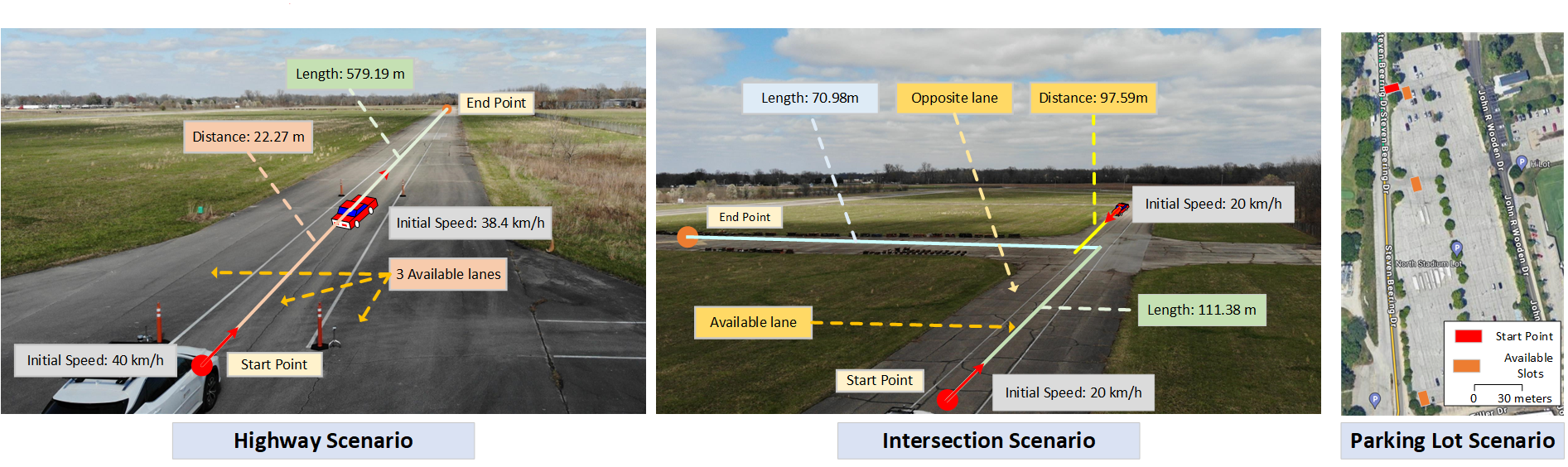}
    \caption{The overview visualization and statistics of the test scenarios.}
    \vspace{-3mm}
    \label{fig:mapstat}
\end{figure*}

\subsection{Execution and Actuator Response} The generated LMPs ($P$) are then sent back from the cloud to the vehicle's ECU, where they are executed. Additionally, we also set two kinds of safety checks for the generated LMPs. First, we will check the format of the LMPs, and if the codes in LMPs are not in a valid format, our \textit{Talk2Drive} framework will not provide any response or action taken in relation to the generated LMPs. Another safety check is parameter verification. It evaluates whether the given parameters are appropriate and safe for the current situation, and prevents the execution of LMPs that could potentially be dangerous. For instance, if the generated LMPs set a target speed that is over the speed limit, our system will disallow the execution of the LMPs. The execution involves adjusting basic driving behaviors and various parameters in the vehicle’s planning and control systems. After ECU executes LMPs, the vehicle's actuators control the throttle, brakes, gear selection, and steering to realize the driving behavior specified by the LMPs through the CAN bus and drive-by-wire system.



\subsection{Memory Module and Personalization}
\label{sec:mm}

In this step, a novel memory module is proposed to store the historical interactions ($H$) between humans and vehicles, which is the key feature of the \textit{Talk2Drive} framework, facilitating its emphasis on personalization. Each interaction between the human and the vehicle is recorded and saved into a memory module in a text format within the ECU. This record includes the humans' commands $I$, LLMs generated LMP $P$, and the human's feedback $F$. This historical data in the memory module is updated after every trip.  The example contents in the memory module are shown in Eq. \ref{eq:mm}:

\begin{align}
&H \to \label{eq:mm} \\
&\left\{\begin{array}{l}
\text{Apart from the requirements I provided before. Here I}\\
\text{will provide you with the history dialogues between the}\\
\text{driver and the vehicle. You will need to learn what}\\
\text{drivers' wants and needs are.}\\
\text{For example, if the history driver's command is "You}\\
\text{drive too conservatively."}\\
\text{The history output action is : ...} \\
\text{After the trip, the driver's feedback is 'A little bit too}\\
\text{fast for me.'}\\
\text{Then the next time your action should adjust according}\\
\text{to the driver's feedback, which is ...}\\
\vdots\\
\text{The history command, action, and driver's feedback are:} \\
\text{Command: I'm on my way to the urgent care.}\\
\text{Action: ...}\\
\text{Evaluation: A little bit too fast.}\\
\vdots
\end{array}\right. \nonumber
\end{align}

Given the adaptive nature of LLMs, if users respond differently to similar commands, the LLMs will prioritize the most recent response as a reference point for their current decision-making process. When a command from the users is issued, the LLMs access the memory module and take stored information ($H$) as part of the input prompts for the decision-making process. Additionally, each user has their own profile in the memory module, ensuring that our framework can deliver personalized driving strategies for everyone.

\section{Experiment Setup}
\label{sec:exp:setup}
\subsection{Autonomous Vehicle Setup}
As shown in Fig. \ref{fig:Vehicle_Setup}, we use an autonomous vehicle to conduct real-world experiments, which is a drive-by-wire enabled 2019 Lexus RX450h with perception sensors, localization module, and communication module. We deploy the open-source autonomous driving software Auoware.AI~\cite{autoware} with ROS Melodic in Ubuntu 18.04. We employ 3D-NDT~\cite{3Dndt} for mapping and localization, and we utilize pure pursuit~\cite{pure_pursuit} for trajectory following.

\begin{figure}[t!]
\centering
\includegraphics[width=\linewidth]{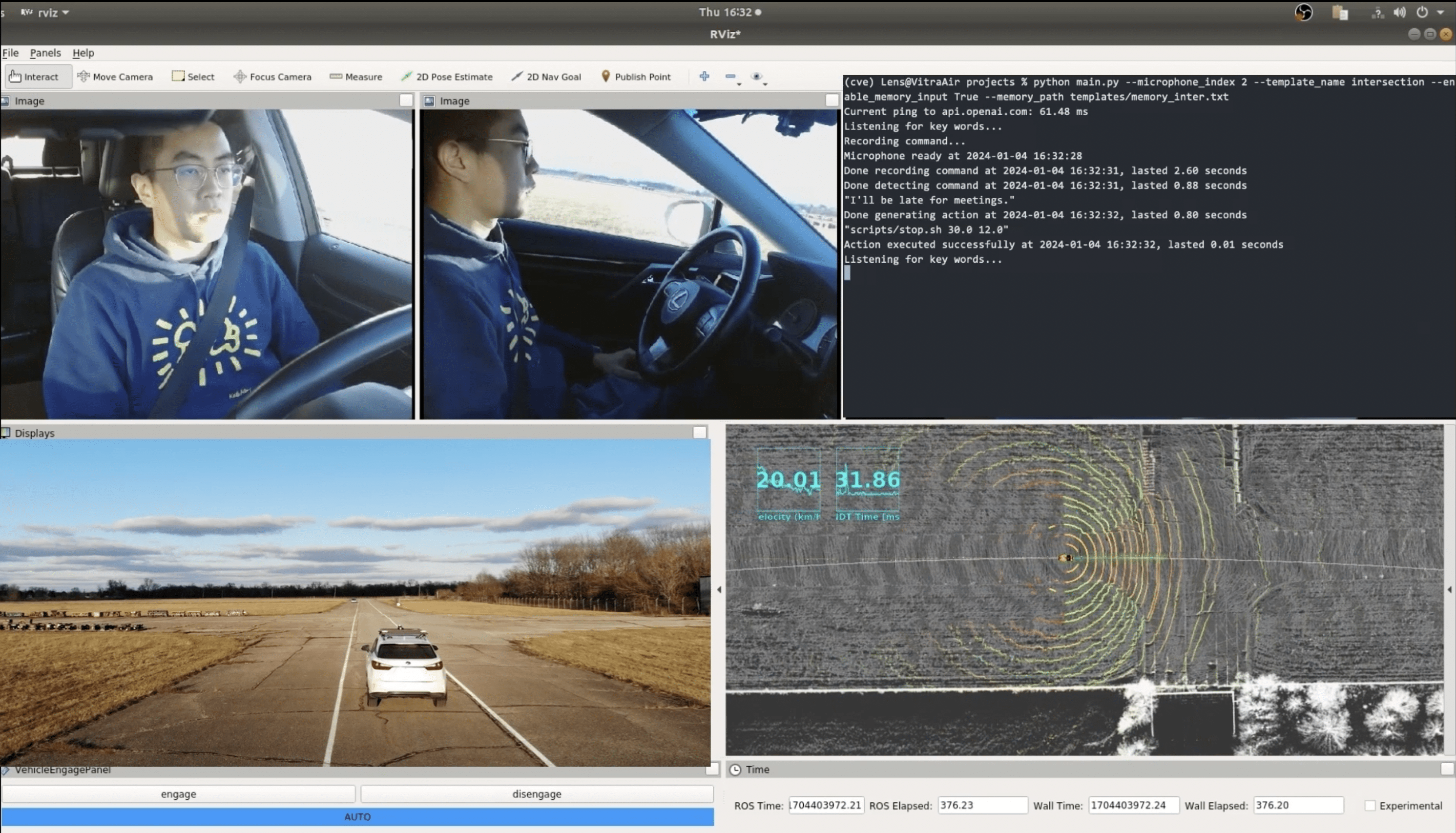}
\caption{The experiment visualization: In the upper left corner is the in-cabin view, while the lower left corner displays the exterior view. The upper right corner shows the console, and the lower right corner presents the lidar map.}
\vspace{-5mm}
\label{fig:exp_vis}
\end{figure}

\subsection{Test Track and Participants}

Our experimental trials include three different scenarios: highway, intersection, and parking lot. \footnote{The experiments conducted in this study satisfy all local traffic guidelines and guarantee the safety of the participants. A human always sits in the driver's seat of the autonomous vehicle to monitor its status and get ready to take over.} The field experiments for the highway and intersection scenarios are conducted at a proving ground in Columbus, IN, USA, aimed at validating the efficacy of the proposed \textit{Talk2Drive} framework. The highway scenario involves a three-way highway, while the intersection contains a two-way junction. Additionally, the parking lot scenario is evaluated on a test track situated at the North Stadium Parking Lot in West Lafayette, IN, USA. The overview visualization and statistics of the test tracks are shown in \ref{fig:mapstat}. The visualization for the experiments is shown in Fig. \ref{fig:exp_vis}.



In our study, we recruit a diverse range of drivers, including individuals of different genders (61.4\% male, 28.6\% female), ages (mean=26.71, std=4.11), and driving experiences(mean=6.79, std=5.08). All participants have valid driving licenses.

\subsection{Input Instructions} \label{subsec:inputinstruction}

In the field of linguistics, giving instructions can be categorized as making a request, which falls under the division of directives, one of the five speech act types according to the general speech act classification theory~\cite{yule2022study}. More specifically, the scale of directness in requests can be characterized by three distinct strategies~\cite{shoshana1989cross}:

\textbf{Direct Strategies} In this scenario, the human explicitly states the desired action, such as “Increase the speed of the vehicle.” This is usually in the form of imperative sentences.

\textbf{Conventionally Indirect Strategies} This scenario involves phrasing requests in a manner that is socially and culturally acknowledged as polite and/or standard. An example would be “Could you please speed up a little bit?”

\begin{table*}[!t]
\centering
\caption{Level of Command Directness and Examples}
\resizebox{1.9\columnwidth}{!}{
\begin{tabular}{c|c|c}
    \toprule
    Command Level & Linguistic Category & Example Commands \\
    \midrule
    \multirow{1}{*}{Level \uppercase\expandafter{\romannumeral1}} & \multirow{1}{*}{Direct and conventionally indirect commands} & Drive as fast as you can. \\
    \midrule
    \multirow{1}{*}{Level \uppercase\expandafter{\romannumeral2}} & \multirow{1}{*}{Non-conventionally indirect commands with strong hints} & You are driving too conservatively.\\
    \midrule
    \multirow{1}{*}{Level \uppercase\expandafter{\romannumeral3}} & \multirow{1}{*}{Non-conventionally indirect commands with mild hints} &  I feel a bit motion-sick right now. \\
    \bottomrule
\end{tabular}
}
\label{tab:directlevel}
\end{table*}
\textbf{Non-Conventionally Indirect Strategies} Requests in this scenario are more implicit and rely on contextual understanding. Within this category, hints can be further divided into strong and mild hints. In the context of requesting speed changes in an autonomous vehicle, strong hints may include explicit comments on the current speed, such as ``You are driving too aggressively.” Conversely, mild hints might be more oblique references to time or urgency, for example, ``I hope we’re not late for the meeting.”

Given that conventionally indirect strategies primarily modify the politeness level of command of direct strategies, we instead divide the non-conventionally indirect requests into two categories based on the strength of the hints. We define our levels of directness in Tab. \ref{tab:directlevel} as our way to classify varying degrees of implicitness in the commands. To gather data, we request our test drivers to generate commands based on their normal speaking preferences and subsequently categorize their commands into our defined levels. Examples of the commands generated are also presented in Tab. \ref{tab:directlevel}.

\subsection{Evaluation Metrics} 
\label{subsec:evalationmetrics}
Our evaluation framework for autonomous vehicles includes driving performance, time efficiency, and personalization. We analyze \textit{Talk2Drive}'s driving performance in terms of safety, using Time to Collision ($\tau$) and speed variance ($\sigma^2$), and comfort, using mean absolute acceleration ($|\bar A|$) and mean absolute jerk ($|\bar J|$). Time efficiency is measured by the LLM latency ($L$), while human satisfaction with personalization is assessed using the takeover rate ($R$).

\textbf{Driving Performance} 
The overall driving performance of \textit{Talk2Drive} is reflected by a driving score ($S$), which is a weighted sum of four sub-scores: Time to Collision score ($S_{\tau}$), speed variance ($S_{\sigma}$), mean absolute acceleration ($S_{|\bar A|}$), and mean absolute jerk ($S_{|\bar J|}$).

\begin{equation}
S = \sum{w_k \cdot S_k}
\end{equation}

For Time to Collision $\tau$, the critical threshold $\tau_c$ is based on human reaction time to take over and brake. Therefore, it can be considered as a hard threshold where any value greater than $\tau_c$ has a score of zero.
\begin{equation}
    S_{\tau} = 
        \begin{cases}
        0,  & \text{if  } \tau_{min} < \tau_c \\
        100, & \text{if  } \tau_{min} \geq \tau_c
        \end{cases}
\end{equation}
where $\tau_c$ is the critical threshold. We chose the value of 1.5 based on existing research results on Time to Collision value \cite{van1994time}. For other metrics like speed variance, acceleration, and jerk, the thresholds are related to individual perception and would vary in different driving scenarios. Therefore, we define the score to quantify the corresponding performance, with a higher score indicating better performance relative to the baseline value. For example, the sub-score for mean absolute jerk $|\bar J|$ can be denoted as:
\begin{equation}
    S_{|\bar J|} = 100 - \gamma\cdot\frac{|\bar J|}{|\bar J|_{\text{baseline}}}
\end{equation}
 where $\gamma$ is the sensitivity factor which is set empirically. The sub-scores for mean absolute acceleration and speed variance are obtained in the same way.

\textbf{LLMs' Latency for Time Efficiency} 
Latency $L$ measures the response time of the LLM, which is important for time-sensitive vehicle applications. To measure latency, we calculate the time difference between the moment a command is sent to the LLMs and when the LLMs return the LMPs:

\begin{equation}
    L = t_{\text{response}} - t_{\text{command}}
\end{equation}
where  $t_{\text{response}}$ is the moment the command is sent to the cloud and $t_{\text{command}}$ is the moment the LMPs are returned to the vehicle.

\textbf{Takeover Rate for Personalization}
The frequency of manual interventions indicates the model's ability to adapt to personalized preferences from different humans~\cite{wang2022gaussian}. The takeover rate $R$ can be calculated as:
\begin{equation}
    R = \frac{N_{\text{takeover}}}{N_{\text{operation}}}
\end{equation}

\noindent where $N_{\text{takeover}}$ is the number of experimental driving trials for drivers involving takeovers while $N_{\text{operation}}$ is the total number of conducted experimental driving trials for them.

\section{Experiment Results}
In the experiments, all participants are required to issue verbal commands to the systems, which will be either baselines or \textit{Talk2Drive} systems chosen randomly. The participants will be unaware of which system is in use. They can decide when to take over the driving system, and the takeover rate will be recorded. Takeovers in this context are noted when humans find the current autonomous system unsatisfactory. Additionally, all driving data will be logged, and the ping latency for our experiments at 200$\sim$400 ms.

\begin{table*}[t!]
\caption{Driving Performance Validation. $\downarrow$: Lower values are better. $\uparrow$: Higher values are better.}
\label{tab:drivingperformance}
\centering
\resizebox{2\columnwidth}{!}{%
\begin{tabular}{c|c|c|cc|cc|c|c}
\toprule
\multirow{3}{*}{\begin{tabular}[c]{@{}c@{}}Driving \\ Secnario\end{tabular}} & \multirow{3}{*}{\begin{tabular}[c]{@{}c@{}}Expected Driving \\ Behavior\end{tabular}} & \multirow{3}{*}{\begin{tabular}[c]{@{}c@{}}Data \\ Type\end{tabular}} & \multicolumn{2}{c|}{Safety Metrics} & \multicolumn{2}{c|}{Comfort Metrics} & Time Efficiency & \multirow{3}{*}{Driving Score$\uparrow$} \\ \cline{4-8}
&  &  & \begin{tabular}[c]{@{}c@{}}Time to \\ Collision($s$)$\uparrow$\end{tabular} & \multicolumn{1}{c|}{\begin{tabular}[c]{@{}c@{}}Speed \\ Variance($m^2/s^2$)$\downarrow$\end{tabular}} & \begin{tabular}[c]{@{}c@{}}Mean Absolute\\ Acceleration($m/s^2$)$\downarrow$\end{tabular} & \multicolumn{1}{c|}{\begin{tabular}[c]{@{}c@{}}Mean Absolute\\ Jerk($m/s^3$)$\downarrow$\end{tabular}} & \begin{tabular}[c]{@{}c@{}}LLM \\ Latency($s$)$\downarrow$\end{tabular} &  \\
\midrule
\multirow{6}{*}{Highway} & \multirow{2}{*}{Overtake} & \textit{Talk2Drive} & 2.63 & 3.44 & 0.22 & 2.69 & 1.80 & \textcolor{green}{87.43} \\
 &  & Baseline & 3.26 & 2.91 & 0.35 & 2.83 & - & 86.00 \\ \cline{2-9}
 & \multirow{2}{*}{Following} & \textit{Talk2Drive} & 7.05 & 1.14 & 0.13 & 2.35 & 1.63 & \textcolor{green}{86.53} \\
 &  & Baseline & 4.02 & 0.78 & 0.22 & 2.50 & - & 86.00 \\ \cline{2-9}
 & \multirow{2}{*}{Right Lane} & \textit{Talk2Drive} & 6.87 & 1.03 & 0.15 & 2.48 & 1.45 & \textcolor{green}{91.52} \\
 &  & Baseline & 4.70 & 7.39 & 0.22 & 2.77 & - & 86.00 \\
\midrule
\multirow{4}{*}{Intersection} & \multirow{2}{*}{Not Yield} & \textit{Talk2Drive} & 0.94 & 0.24 & 0.27 & 2.38 & 1.65 & \textcolor{green}{59.87} \\
 &  & Baseline & 1.14 & 0.46 & 0.46 & 2.34 & - & 56.00 \\ \cline{2-9}
 & \multirow{2}{*}{Yield} & \textit{Talk2Drive} & - & 0.24 & 0.61 & 2.36 & 1.43 & \textcolor{green}{90.98} \\
 &  & Baseline & - & 1.67 & 0.90 & 2.32 & - & 86.00 \\
\bottomrule
\end{tabular}%
}
\end{table*}

\begin{table}[t!]
\caption{Takeover Rate Improvement through Personalization. }
\label{tab:personalization}
\centering
\resizebox{\columnwidth}{!}{%
\begin{tabular}{c|cccc}
\toprule
\begin{tabular}[c]{@{}c@{}}Driving\\ Scenario\end{tabular} & \begin{tabular}[c]{@{}c@{}}Command \\ Directness\end{tabular} & Baseline & \textit{Talk2Drive} & Reduction \\
\midrule
\multirow{3}{*}{Highway} & I & 0.33 & 0.07 & 78.8\% \\
 & II & 0.63 & 0.20 & 68.3\% \\
 & III & 0.77 & 0.31 & 59.7\% \\
\midrule
\multirow{3}{*}{Intersection} & I & 0.33 & 0.11 & 66.7\% \\
 & II & 0.71 & 0.29 & 59.2\% \\
 & III & 0.48 & 0.21 & 56.3\% \\
\midrule
\multirow{3}{*}{Parking} & I & 0.07 & 0 & 100\% \\
 & II & 0.20 & 0 & 100\% \\
 & III & 0.67 & 0.24 & 64.2\% \\
\bottomrule
\end{tabular}%
}
\end{table}

\subsection{The Validation of the Driving Performance}

To validate the driving performance of the \textit{Talk2Drive} system, we conduct comprehensive experiments to compare the driving performances between the baseline and the \textit{Talk2Drive} through the metrics defined in Sec. \ref{subsec:evalationmetrics}. The baseline values are from the average data of human drivers. We categorized the collected data based on driving behaviors: in highway scenarios, data is classified into following the front vehicle and staying in the current lane, overtaking by changing to the left lane, and changing to the right lane; whereas in intersection scenarios, data is categorized into yielding to approaching vehicles and not yielding. The average driving performance for each scenario is shown in Tab. \ref{tab:drivingperformance}. Note that for each expected driving behavior the driving pattern is distinct, therefore a baseline is created for every driving behavior separately.


\textbf{Safety} Time to Collision (TTC) and speed variance are used to evaluate the safety of the Talk2Drive system. For highway scenarios, TTCs with Talk2Drive enabled are above the 1.5-second safety threshold (the driver will have enough time to react to avoid rear-end collisions~\cite{van1994time}) and comparable to the baseline. In intersection scenarios, TTCs for not yielding cases are greater than the threshold, which is reasonable since not yielding is an aggressive behavior in itself. For yielding cases, TTC is not applicable since the two vehicles are always in different lanes. Speed variance is similar to the baseline in highway scenarios and lower in intersection scenarios, indicating that Talk2Drive ensures a steady and consistent drive.

\textbf{Comfort} Sudden acceleration and deceleration, or frequent change in velocity are two major causes of discomfort during vehicle operation. Several studies on vehicle comfort indicate an acceleration less than or equal to $0.56 m \cdot s^{-2}$ can be considered a ``Very Good" ride experience, while a jerk less than $2.94 m/s^3$ is acceptable to human~\cite{de2023standards,hoberock1976survey}.
Tab.~\ref{tab:drivingperformance} reveals that mean acceleration and jerk do not substantially exceed the baseline levels while also not exceeding the suggested threshold for the best riding experience. Such results show the speed adjustments made through \textit{Talk2Drive} ensure the same level of riding comfort as human drivers.

\begin{table}[!t]
\centering
\caption{Effectiveness of Memory Modules in Influencing Takeover Rates}
\begin{tabular*}{0.9\linewidth}{@{\extracolsep{\fill}}c|cc@{\extracolsep{\fill}}}
    \toprule
     \makecell{Driver } &  \makecell{Without Memory Module} & \makecell{With Memory Module}  \\
    \midrule
    A & 0.14  & 0.07 (50.0\%$\downarrow$)\\
    B & 0.23  & 0.08 (65.2\%$\downarrow$) \\
    C & 0.29  & 0.18 (37.9\%$\downarrow$)\\
    \bottomrule
\end{tabular*}
\label{tab:takeoverrate}
\end{table}

\textbf{Time Efficiency} Given that the latency of the speech-to-text module remains mostly consistent (around 300ms) under the same network condition, we focus on the duration from the initiation of the LLM API call to the successful reception of the command text. As is shown in the table, the latency of GPT-4~\cite{openai_gpt-4_2023} remains stable at around 1.6 seconds, which is acceptable in non-urgent scenarios.

\textbf{Driving Score} As defined in Sec. \ref{subsec:evalationmetrics}, the driving score is the weighted sum of all the safety and comfort metrics. Tab. \ref{tab:drivingperformance} presents the calculated driving scores, with green text highlighting scores that exceed the baseline for the corresponding driving behavior. As shown in the table, all the driving scores are higher than the baseline, indicating that our approach offers enhanced comfort and safety compared to human drivers.


\subsection{The Improvement on Personalization}

This section explores the impact of integrating the \textit{Talk2Drive} framework into autonomous driving systems to enhance personalization based on the LLM model GPT-4~\cite{openai_gpt-4_2023}. One of the central focuses of our system is its ability to offer drivers a personalized driving experience. Through \textit{Talk2Drive}, individuals can express their preferences or feelings with varying degrees of directness, prompting the system to make corresponding adjustments. The commands collected are divided into three levels of directness, as explained in Tab. \ref{tab:directlevel}. 

For each command directness within each expected driving behavior, the system's responses to various responses given by human participants are collected and calculated into the defined metrics.  Here, we utilize the takeover rate as our primary metric, and we assess the takeover rate both pre- and post-integration of \textit{Talk2Drive}. We incorporate a rule-based system with a keyword-trigger logic as the baseline for comparison. For instance, in the highway scenario, commands like ``accelerate," ``left," and ``right" prompt the system to execute relevant actions such as speeding up or changing lanes accordingly.  Results are gathered in diverse driving scenarios on human drivers with diverse driving styles. 

As shown in Tab. \ref{tab:personalization}, the takeover rate in all driving scenarios has decreased significantly, ranging from a 56.3\% reduction to the complete elimination of takeover cases, demonstrating the \textit{Talk2Drive} system's ability to personalize the driving experience based on human preference. Additionally, one important finding from our results is that for all levels of command directness, our \textit{Talk2Drive} system shows significant improvements compared to the baseline systems. The \textit{Talk2Drive} system performs very well for all three levels of command directness, with even the highest takeover rate being only 0.31.



\subsection{The Effectiveness of the Memory Module}


To further investigate the performance of our memory module, we also conducted experiments comparing personalization across two settings: conditions without the memory module, and conditions with the memory module. These experiments took place in the parking lot, where the vehicle drives through various autonomous settings, including speed adjustments based on human inputs. Humans take over in this context when they find the adjusted speed unsatisfactory. We conducted performance comparisons across these settings for three different drivers. As demonstrated in Tab. \ref{tab:takeoverrate}, there are significant reductions in takeover rate with \textit{Talk2Drive}, regardless of the driver's aggression or conservatism levels. 

The results in Tab. \ref{tab:takeoverrate} reveal that the inclusion of the memory module leads to a marked reduction in the takeover rate. The utilization of \textit{Talk2Drive} framework without the memory module brings takeover rates between 0.14 and 0.29. When implementing \textit{Talk2Drive} framework with the memory module, the takeover rate can further decrease up to only 0.07. Compared to their personalization performance, Talk2Drive with a memory module can reduce takeover rates by up to 65.2\% compared to those without the memory module, which illustrates the benefits of a history-recording module in achieving a more personalized driving experience.

\section{Conclusions}
\label{sec:concl}

In this paper, we proposed an LLM-based framework, \textit{Talk2Drive}, to translate natural verbal commands into executable controls for autonomous vehicles and learn to satisfy personal preferences for safety, efficiency, and comfort with a proposed memory module. Real-world experiments proved that the proposed system can comprehend human intentions at different intuition levels, ranging from direct commands like ``can you drive faster'' to indirect commands like ``I am really in a hurry now''. We adopted the takeover rate to quantify the trust of human drivers in the proposed system, where \textit{Talk2Drive} was shown to significantly reduce the takeover rate in highway, intersection, and parking scenarios. We also validated that the proposed memory module can consider personalized preferences and further reduce the takeover rate by up to 65.2\% compared with those without a memory module. In our future research on the Talk2Drive framework, we will focus on reducing LLMs' latency with technologies like model distillation to achieve real-time performance, aligning with the stringent requirements for rapid interaction.



\bibliography{bib/previous,bib/previous2,bib/can,bib/zichong}
\bibliographystyle{IEEEtran}

\end{document}